\documentclass[runningheads]{llncs}
\usepackage[T1]{fontenc}
\usepackage{graphicx}
\usepackage{hyperref}
\usepackage{orcidlink}
\usepackage{booktabs}
\usepackage{multirow}
\usepackage{pifont} 
\usepackage{array}
\begin{document}
\title{P2MFDS: A Privacy-Preserving Multimodal Fall Detection System for Elderly People in Bathroom Environments}
\titlerunning{P2MFDS: A Multimodal Fall Detection System}
%
%
\author{Haitian Wang\orcidlink{0009-0000-7544-4667}\inst{1,2} \and
Yiren Wang\orcidlink{0009-0009-7435-3398}\inst{2} \and
Xinyu Wang\orcidlink{0009-0008-4065-2472}\inst{2} \and
Yumeng Miao\orcidlink{0009-0007-5204-2201}\inst{2} \and 
Yuliang Zhang\orcidlink{0000-0001-8064-3003}\inst{2} \and
Yu Zhang\orcidlink{0000-0002-3952-670X}\inst{1}\thanks{Corresponding author: \email{zhangyu@nwpu.edu.cn}} \and
Atif Mansoor\orcidlink{0000-0001-6940-3914}\inst{2}}
\authorrunning{H. Wang et al.}
%
\institute{
School of Computer Science, Northwestern Polytechnical University,\\
No. 1 Dongxiang Road, Chang'an District, Xi’an 710129, China
\and
Department of Computer Science and Software Engineering,\\
The University of Western Australia, 35 Stirling Hwy, Crawley, WA 6009, Australia\\
}

\maketitle              
\pagestyle{empty}
\begin{abstract}
By 2050, people aged 65 and over are projected to make up 16\% of the global population. As aging is closely associated with increased fall risk, particularly in wet and confined environments such as bathrooms where over 80\% of falls occur. Although recent research has increasingly focused on non-intrusive, privacy-preserving approaches that do not rely on wearable devices or video-based monitoring, these efforts have not fully overcome the limitations of existing unimodal systems (e.g., WiFi-, infrared-, or mmWave-based), which are prone to reduced accuracy in complex environments. These limitations stem from fundamental constraints in unimodal sensing, including system bias and environmental interference, such as multipath fading in WiFi-based systems and drastic temperature changes in infrared-based methods. To address these challenges, we propose a Privacy-Preserving Multimodal Fall Detection System for Elderly People in Bathroom Environments. First, we develop a sensor evaluation framework to select and fuse millimeter-wave radar with 3D vibration sensing, and use it to construct and preprocess a large-scale, privacy-preserving multimodal dataset in real bathroom settings, which will be released upon publication. Second, we introduce P2MFDS, a dual-stream network combining a CNN–BiLSTM–Attention branch for radar motion dynamics with a multi-scale CNN–SEBlock–Self-Attention branch for vibration impact detection. By uniting macro- and micro-scale features, P2MFDS delivers significant gains in accuracy and recall over state-of-the-art approaches. Code and pretrained models are available at \href{https://github.com/HaitianWang/P2MFDS-A-Privacy-Preserving-Multimodal-Fall-Detection-Network-for-Elderly-Individuals-in-Bathroom}{P2MFDS Github Repository}.

\end{abstract}

\vspace{-4mm}
\section{Introduction}
\vspace{-4mm}

By 2050, people aged 65 and above are projected to constitute 16\% of the global population, up from 10\% in 2022\cite{leong2018world}. A growing number of elderly people prefer to live independently, yet this increases their vulnerability to medical emergencies such as falls, which remain one of the leading causes of injury-related morbidity and mortality among elderly adults\cite{strini2021fall}\cite{wang2020elderly}. Bathrooms, with their slippery surfaces, are particularly hazardous; studies indicate that over 80\% of elderly falls occur in these environments\cite{singh2020sensor}. Given that falls can cause severe injuries or deaths, a real-time and accurate solution to detect falls in the bathroom is essential for the elderly \cite{yazar2014multi}\cite{akash2023elderly}.

Existing fall detection solutions for elderly people can be categorized into three primary approaches. The first involves in-home caregivers who provide continuous monitoring and immediate assistance in the event of a fall \cite{chen2020fall}\cite{tun2021internet}. Although effective, this solution is cost-prohibitive and lacks scalability\cite{cardoso2020care}. The second approach utilizes wearable devices, such as smart watches and electrocardiogram (ECG) monitors. These devices continuously track physiological parameters such as heart rate, blood pressure, and vibration\cite{guinon2007moving}. However, these devices face adoption challenges, particularly among elderly users who may find them uncomfortable, forget to wear them, or struggle to use them in wet environments such as bathrooms\cite{perumal2024review}. The third approach integrates smart home technologies, including video-based and audio-based fall detection systems\cite{palipana2018falldefi} \cite{riquelme2019ehomeseniors}. While these systems can achieve high detection accuracy, their deployment in private spaces, such as bathrooms, raises significant privacy concerns, limiting user acceptance. Given these limitations, there is a critical need for a non-intrusive, real-time fall detection system that ensures both accuracy and privacy in bathroom environments\cite{wagner2022non}.

Recent research has explored non-intrusive fall detection methods, focusing on privacy-preserving solutions that do not require wearable devices or video-based monitoring \cite{mavaddat2014millimeter}\cite{gerstmair2018highly}. Palipana\cite{palipana2018falldefi} et al. proposed FallDeFi, a WiFi-based system that detects falls by analyzing signal disturbances. While effective in controlled environments, its accuracy degrades in complex spaces like bathrooms due to multipath interference and environmental noise. Similarly, Zigel et al.\cite{zigel2009method} developed a vibration- and sound-based detection system that addresses privacy concerns but is highly sensitive to background noise, leading to false positives from non-fall events such as dropped objects. Akash et al.\cite{akash2023elderly} explored millimeter-wave (mmWave) radar for fall detection, utilizing its high-resolution motion tracking capabilities. 

\begin{figure*}[tb]
    \centering
    \includegraphics[width=1\linewidth]{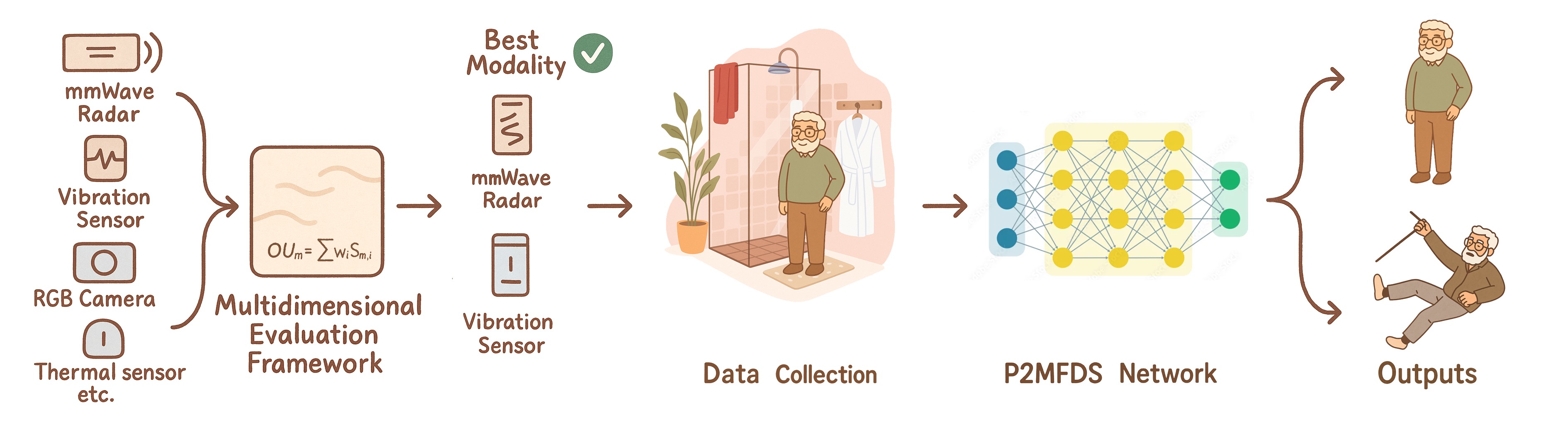}
    \vspace{-8mm}
    \caption{System overview of the P2MFDS pipeline. First, a multidimensional evaluation framework scores candidate sensing modalities (e.g., mmWave radar, vibration sensor, RGB and thermal cameras) to select the optimal combination. Next, multimodal data are collected in a realistic bathroom setup. The P2MFDS Network then fuses long‐term motion features extracted by a 1D CNN–BiLSTM–Attention stream with short‐term impact signatures captured by a Multi‐Scale CNN–SEBlock–Self‐Attention stream. Finally, the fused representation is classified into fall or non‐fall events.}
    \label{fig:overview}
    \vspace{-8mm}
\end{figure*}

However, its effectiveness depends on precise calibration and can be influenced by bathroom wall tiles and flooring \cite{neipp2003analysis}\cite{taha2021intelligent}. 
These approaches predominantly rely on unimodal sensor data, which undergoes preprocessing, feature extraction, and classification to detect falls \cite{chen2021fall}\cite{hashim2020accurate}.However, unimodal systems face fundamental limitations: the performance is approaching theoretical bounds, leaving minimal room for improvement through feature engineering alone \cite{appeadu2023falls}\cite{too2017classification}. Moreover, single-sensor data is highly susceptible to system bias and environmental interference. For instance, infrared sensors are affected by temperature variations, while vibration sensors can misinterpret background noise or minor disturbances as falls, leading to increased false alarm rates. These challenges underscore the need for a multimodal, sensor-fusion approach to enhance accuracy and robustness in real-world applications \cite{khan2020wireless}\cite{kim2020systematic}.

\begin{table}[!b]
\vspace{-6mm}
\caption{Proposed evaluation framework for selecting optimal sensing modalities in privacy-preserving fall detection.}
\label{tab:selection_criteria}
\centering
\setlength{\tabcolsep}{3pt}
\begin{tabular}{
    p{3.5cm} 
    p{5.2cm} 
    p{1.8cm} 
    >{\centering\arraybackslash}p{1.2cm}
}
\toprule
\textbf{Selection Criteria} & \textbf{Description} & \textbf{Rating} & \textbf{Weight} \\
\midrule
Target Relevance      & Relevance for fall detection & 1--3 scale   & 3 \\
Non-Intrusiveness      & Level of privacy preservation & 1--5 scale   & 2 \\
Energy Efficiency     & Power consumption (AC/DC) & 1--5 scale   & 3 \\
Comp Complexity       & Processing demands & 1--5 scale   & 2 \\
Deployability         & Ease of integration & 1--5 scale   & 2 \\
Recall Rate           & Ability to detect falls & 1--5 scale   & 3 \\
Availability          & Reliability in real conditions & 1--5 scale   & 2 \\
Cost-Effectiveness    & Economic feasibility & 1--5 scale   & 1 \\
\bottomrule
\end{tabular}
\vspace{-8mm}
\end{table}

\begin{table}[!t]
\vspace{-6mm}
\caption{
Evaluation of 14 sensing modalities across nine key factors for privacy-preserving fall detection in bathrooms. TR: Target Relevance, CE: Cost Effectiveness, Rec: Recall, EE: Energy Efficiency, CC: Computational Complexity, Dep: Deployability, NI: Non-Intrusiveness, Avail: Availability, Score: Usability Score.
}
\label{tab:detection_methods}
\centering
\setlength{\tabcolsep}{3pt}
\begin{tabular}{
    p{2.6cm} 
    >{\centering\arraybackslash}p{0.8cm}
    >{\centering\arraybackslash}p{0.8cm}
    >{\centering\arraybackslash}p{0.8cm}
    >{\centering\arraybackslash}p{1.2cm}
    >{\centering\arraybackslash}p{0.8cm}
    >{\centering\arraybackslash}p{0.8cm}
    >{\centering\arraybackslash}p{0.8cm}
    >{\centering\arraybackslash}p{0.8cm}
    >{\centering\arraybackslash}p{0.9cm}
}
\toprule
\textbf{Method} & \textbf{TR} & \textbf{CE} & \textbf{Rec} & \textbf{EE} & \textbf{CC} & \textbf{Dep} & \textbf{NI} & \textbf{Avail} & \textbf{Score} \\
\midrule
Wi-Fi & 3 & 3 & 4 & 2 & 4 & 3 & 5 & 3 & 61 \\
Infrared & 3 & 2 & 3 & 3 & 5 & 4 & 5 & 1 & 61 \\
Bluetooth & 3 & 4 & 4 & 2 & 4 & 3 & 5 & 2 & 64 \\
\textbf{MMW} & 3 & 3 & 5 & 4 & 5 & 4 & 5 & 3 & \textbf{73} \\
Ultrasonic & 3 & 3 & 4 & 2 & 4 & 4 & 4 & 3 & 62 \\
Audio & 3 & 4 & 4 & 3 & 3 & 4 & 2 & 4 & 65 \\
Thermal Imaging & 3 & 1 & 5 & 5 & 3 & 3 & 4 & 1 & 62 \\
\textbf{3D Vibration} & 3 & 4 & 4 & 4 & 3 & 3 & 5 & 4 & \textbf{68} \\
Light Sensors & 1 & 5 & 4 & 1 & 2 & 1 & 5 & 5 & 47 \\
Door Sensors & 1 & 5 & 4 & 2 & 2 & 5 & 5 & 4 & 62 \\
Furniture Sensors & 1 & 5 & 2 & 1 & 1 & 5 & 5 & 3 & 51 \\
Weight Sensors & 1 & 5 & 3 & 1 & 1 & 5 & 5 & 4 & 57 \\
Electricity Usage & 1 & 2 & 1 & 1 & 2 & 2 & 5 & 4 & 63 \\
Water Usage & 1 & 2 & 3 & 1 & 2 & 3 & 5 & 4 & 54 \\
\bottomrule
\end{tabular}
\vspace{-8mm}
\end{table}

To solve this issue, we propose a Privacy-Preserving Multimodal Fall Detection System for Elderly People in Bathroom Environments. This system comprises two parts (as shown in Fig.~\ref{fig:overview})). The first part is a systematic sensor evaluation framework that guides the selection and fusion of millimeter-wave radar with 3D vibration sensing, enabling the collection and preprocessing (smoothing, filtering, fusion) of a large-scale, privacy-preserving multimodal dataset in real bathroom settings; this dataset will be released upon publication. The second part is P2MFDS, a dual-stream network combining a CNN–BiLSTM–Attention branch for capturing long-term motion dynamics from radar point clouds and a multi-scale CNN–SEBlock–Self-Attention branch for detecting short-term impact signatures in vibration data. By uniting macro- and micro-scale features, P2MFDS achieves 95.0\% accuracy, 87.9\% recall, and a 91.3\% F1-score, substantially outperforming state-of-the-art unimodal approaches. The complete implementation and pretrained models are available at \href{https://github.com/HaitianWang/P2MFDS-A-Privacy-Preserving-Multimodal-Fall-Detection-Network-for-Elderly-Individuals-in-Bathroom}{P2MFDS Github Repository}.

\vspace{-4mm}
\section{Offline Multimodal Sensors Selection} 
\vspace{-4mm}

To develop a robust non-intrusive fall detection system suitable for bathroom environments, we propose a comprehensive Multimodal Sensor Evaluation Framework (as shown in Table~\ref{tab:selection_criteria}). This framework systematically assesses various sensing modalities based on their feasibility, performance, and deployability in real-world conditions. The evaluation criteria include accuracy, non-intrusiveness, energy efficiency, computational complexity, deployability, recall rate, availability, and cost-effectiveness.

Each sensing modality is assigned a weighted score for the eight evaluation dimensions, contributing to an overall usability score (\(OU\)) using the following equation: \(OU_m = \sum_{i=1}^{N} w_i \times S_{m,i}\), where \(OU_m\) represents the overall usability of modality \(m\), \(S_{m,i}\) denotes the score for modality \(m\) in the \(i\)-th dimension, and \(w_i\) is the weight assigned to the \(i\)-th evaluation criterion. Higher \(OU_m\) values indicate superior suitability for bathroom fall detection.

Using the Multimodal Sensor Evaluation Framework, we evaluate 14 widely used sensing modalities across multiple categories. The evaluation results (as shown in Table~\ref{tab:detection_methods}) indicate that mmWave radar and a three-axis vibration achieve the highest usability scores, making them the most suitable choices for privacy-preserving fall detection in bathroom environments. The mmWave radar offers high-precision motion tracking that remains effective in varying lighting conditions while maintaining privacy, whereas the three-axis vibration provides real-time vibration sensing with low power consumption, ensuring reliable detection. The combination of these two modalities enables a complementary approach, where the mmWave radar captures overall motion patterns, and the vibration sensor detects localized ground vibrations. This multimodal integration enhances detection accuracy, minimizes false positives, and improves robustness against environmental noise, making the system well-suited for real-world deployment.

\vspace{-4mm}
\section{Privacy-Preserving Multimodal Fall Detection System Network}
\label{Privacy-Preserving Multimodal Fall Detection System Network}
\vspace{-4mm}

\begin{figure*}[tb]
    \centering
    \includegraphics[width=\linewidth]{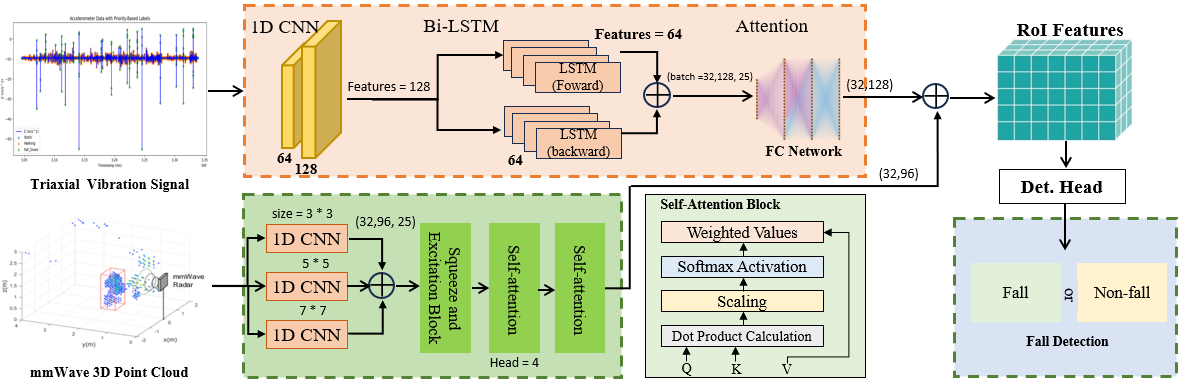}
    \vspace{-8mm}
    \caption{Overview of the P2MFDS network architecture. The upper pipeline employs a 1D CNN--BiLSTM--Attention sequence to extract global motion features from mmWave 3D point clouds, while the lower pipeline utilizes multi-scale CNN, SE block, and self-attention to capture localized vibration signals. Both streams are fused to enable robust, privacy-preserving fall detection in bathroom environments.}
    \label{fig:P2MFDS_Network}
    \vspace{-6mm}
\end{figure*}

Privacy-Preserving Multimodal Fall Detection System (P2MFDS) network fuses mmWave radar and triaxial vibration streams to exploit their complementary sensing characteristics. After targeted low-pass filtering to remove high-frequency noise while preserving fall-related signatures, the radar branch captures macro-scale motion attributes (e.g., velocity, distance, signal energy) and the vibration branch extracts micro-scale impact signatures. Two independent processing streams—one employing a CNN–BiLSTM–Attention pipeline for radar point clouds and the other using a Multi-Scale CNN–SEBlock–Self-Attention pipeline for vibration data—then distill salient features. Finally, these embeddings are concatenated and fed into a lightweight classifier for fall vs.\ non-fall prediction (Fig.~\ref{fig:P2MFDS_Network}).

\vspace{-2mm}
\subsection{Signal Preprocessing}
\vspace{-2mm}

Prior to feature extraction, each modality undergoes targeted low-pass filtering to suppress high-frequency noise while preserving fall-related signatures. For vibration data, a Moving Average Filter (MAF) smooths transient disturbances by computing

\begin{equation}
y_{\mathrm{vib}}(t) = \frac{1}{N} \sum_{i=0}^{N-1} x_{\mathrm{vib}}(t - i),
\label{eq:maf}
\end{equation}

where \(N\) is chosen to balance noise reduction against responsiveness. For radar data, a first-order Exponential Low-Pass Filter (ELPF) adapts to rapid motion changes:

\begin{equation}
y_{\mathrm{radar}}(t) = \alpha\, x_{\mathrm{radar}}(t) + (1 - \alpha)\, y_{\mathrm{radar}}(t - 1),
\label{eq:elpf}
\end{equation}

\vspace{-2mm}

with \(\alpha\in(0,1)\) controlling the trade-off between stability and sensitivity. This dual‐filtering strategy mitigates environmental artifacts (e.g., electrical interference, multipath reflections) and enhances the robustness of subsequent feature extraction.

\vspace{-2mm}
\subsection{3D Vibration-Based Impact Feature Extraction}
\vspace{-2mm}
To capture fine-grained temporal dynamics and transient impacts related to fall events, the triaxial vibration signal initially undergoes feature extraction through a 1D Convolutional Neural Network (1D CNN) and a Bidirectional Long Short-Term Memory (Bi-LSTM) network. Specifically, the input vibration data is first convolved by multiple CNN kernels to produce intermediate temporal-spatial embeddings, which are subsequently fed into a Bi-LSTM composed of forward and backward sequences. The Bi-LSTM effectively integrates long-term dependencies from historical and future contexts, yielding comprehensive temporal embeddings as follows:

\begin{equation}
\overrightarrow{h_t} = \mathrm{LSTM}_{\mathrm{fw}}(X_t, \overrightarrow{h_{t-1}}), \quad
\overleftarrow{h_t} = \mathrm{LSTM}_{\mathrm{bw}}(X_t, \overleftarrow{h_{t+1}})
\end{equation}

where $\overrightarrow{h_t}$ and $\overleftarrow{h_t}$ represent forward and backward hidden states at time step $t$, respectively. These embeddings are concatenated to form the final hidden state $h_t = [\overrightarrow{h_t}, \overleftarrow{h_t}]$, effectively capturing bidirectional temporal information. 

To highlight critical temporal segments indicative of potential falls, an attention mechanism is subsequently applied. Given the Bi-LSTM output $H = \{h_1, h_2, \dots, h_T\}$, the attention weights $\alpha_t$ are calculated by:

\vspace{-6mm}

\begin{equation}
\alpha_t = \frac{\exp(h_t W_a)}{\sum_{i=1}^{T} \exp(h_i W_a)}, \quad
F_{\mathrm{vibration}} = \sum_{t=1}^{T} \alpha_t h_t
\end{equation}

where \( W_a \) denotes learnable parameters, and \( F_{\mathrm{vibration}} \) represents the refined impact feature embedding emphasizing the most informative segments of vibration data.

\vspace{-2mm}
\subsection{mmWave Radar-Based Motion Feature Extraction}
\vspace{-2mm}

The mmWave radar stream focuses on capturing macro-scale motion patterns within the 3D point cloud data through a multi-scale convolutional strategy enhanced by channel-wise and temporal attention mechanisms. Initially, radar signals are processed by parallel 1D CNN branches with varying kernel sizes (3, 5, and 7) to capture diverse motion scales. These multi-scale features are merged into a unified embedding for subsequent recalibration.

\begin{figure*}[!b]
    \centering
    \includegraphics[width=\linewidth]{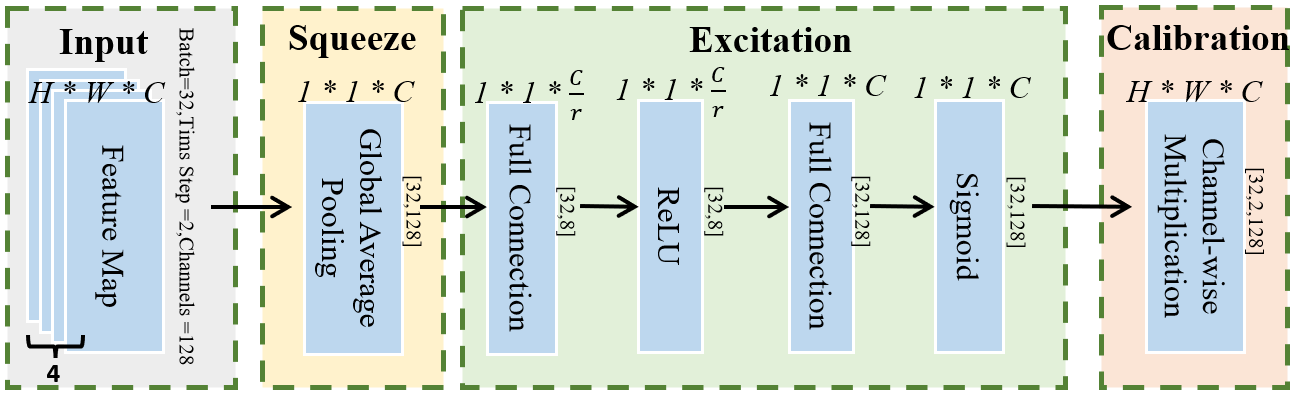}
    \vspace{-8mm}
    \caption{Architecture of the SE block. Channel-wise statistics are extracted via global average pooling, passed through two fully connected layers with a bottleneck structure, and used to generate attention weights via sigmoid activation. The input feature map is then recalibrated through channel-wise multiplication.}
    \label{fig:SEBlock}
    \vspace{-8mm}
\end{figure*}

To further enhance critical channel-level characteristics and suppress redundant features, we incorporate a Squeeze-and-Excitation (SE) block \cite{hu2018squeeze}. the SE block contributes to improved generalization by reducing the risk of overfitting through dynamic channel reweighting, which is particularly beneficial under data-limited training conditions. The SE block first applies global average pooling (GAP) to aggregate global spatial information across each channel, forming a channel descriptor $z_c$:
\vspace{-2mm}
\begin{equation}
z_c = \frac{1}{H \times W} \sum_{i=1}^{H} \sum_{j=1}^{W} x_c(i, j)
\end{equation}

where $x_c(i,j)$ denotes the feature map value at position $(i,j)$ of channel $c$ (as shown in Fig.~\ref{fig:SEBlock}). Subsequently, channel-wise recalibration weights are learned through fully connected layers with a bottleneck architecture:
\begin{equation}
s_c = \sigma\left( W_2\, \mathrm{ReLU}(W_1 z_c) \right), \quad
\hat{x}_c = s_c \cdot x_c
\end{equation}
where $W_1$ and $W_2$ represent learnable weights of fully connected layers, $\sigma$ denotes the sigmoid activation function, and $\hat{x}_c$ is the recalibrated channel-wise feature map. This recalibration adaptively emphasizes significant motion-related features across channels.

Finally, a self-attention mechanism utilizes these recalibrated features to enhance temporal sensitivity and motion dynamics (as shown in Fig.~\ref{fig:P2MFDS_Network}). Query ($Q$), Key ($K$), and Value ($V$) matrices are constructed, and the self-attention output is computed as:
\vspace{-2mm}

\begin{equation}
F_{\mathrm{radar}} = \mathrm{softmax}\left( \frac{Q K^{\mathrm{T}}}{\sqrt{d_k}} \right) V
\end{equation}

Here, \( d_k \) is the dimensionality of keys, and the resulting embedding \( F_{\mathrm{radar}} \) captures the most salient temporal motion characteristics relevant to accurate fall detection.

\vspace{-2mm}
\subsection{Multimodal Feature Fusion and Classification}
\vspace{-2mm}

 After the mmWave 3D point cloud and vibration streams independently extract motion and impact features, respectively, these two embeddings are concatenated to form a joint representation:
\vspace{-3mm}
 {\begin{equation}
F_{\mathrm{fusion}} = \sigma\bigl(W_f [F_{\mathrm{radar}}; F_{\mathrm{acc}}] + b_f\bigr),
\end{equation}}
 {where \( W_f \) and \( b_f \) are learnable parameters, and \(\sigma\) is the activation function. This fused feature map (labeled “RoI Features” in Fig.~\ref{fig:P2MFDS_Network}) is then fed into a lightweight \emph{detection head} (Det Head)—a final classification module comprising fully connected layers and a confidence function. The Det Head transforms the high-level fused representation into discrete predictions (fall vs.\ non-fall)\cite{kiranyaz20211d} \cite{alzubaidi2021review}.}
\vspace{-2mm}

\vspace{-4mm}
\section{Experiment}
\label{sec:Experiment}
\vspace{-4mm}

This section presents the experimental design and setup used to evaluate the proposed P2MFDS Network in real-world bathroom scenarios. We first describe the testing environment and sensor placement, followed by the methodology for collecting multimodal data under various fall and non-fall conditions. Finally, we present the metrics and benchmarking procedures used to validate our system's effectiveness and robustness.

\vspace{-2mm}
\subsection{Experiment Setup}
\label{sec:Experiment_Setup}
\vspace{-2mm}

\begin{figure*}[tb]
  \centering
  \includegraphics[width=0.9\linewidth]{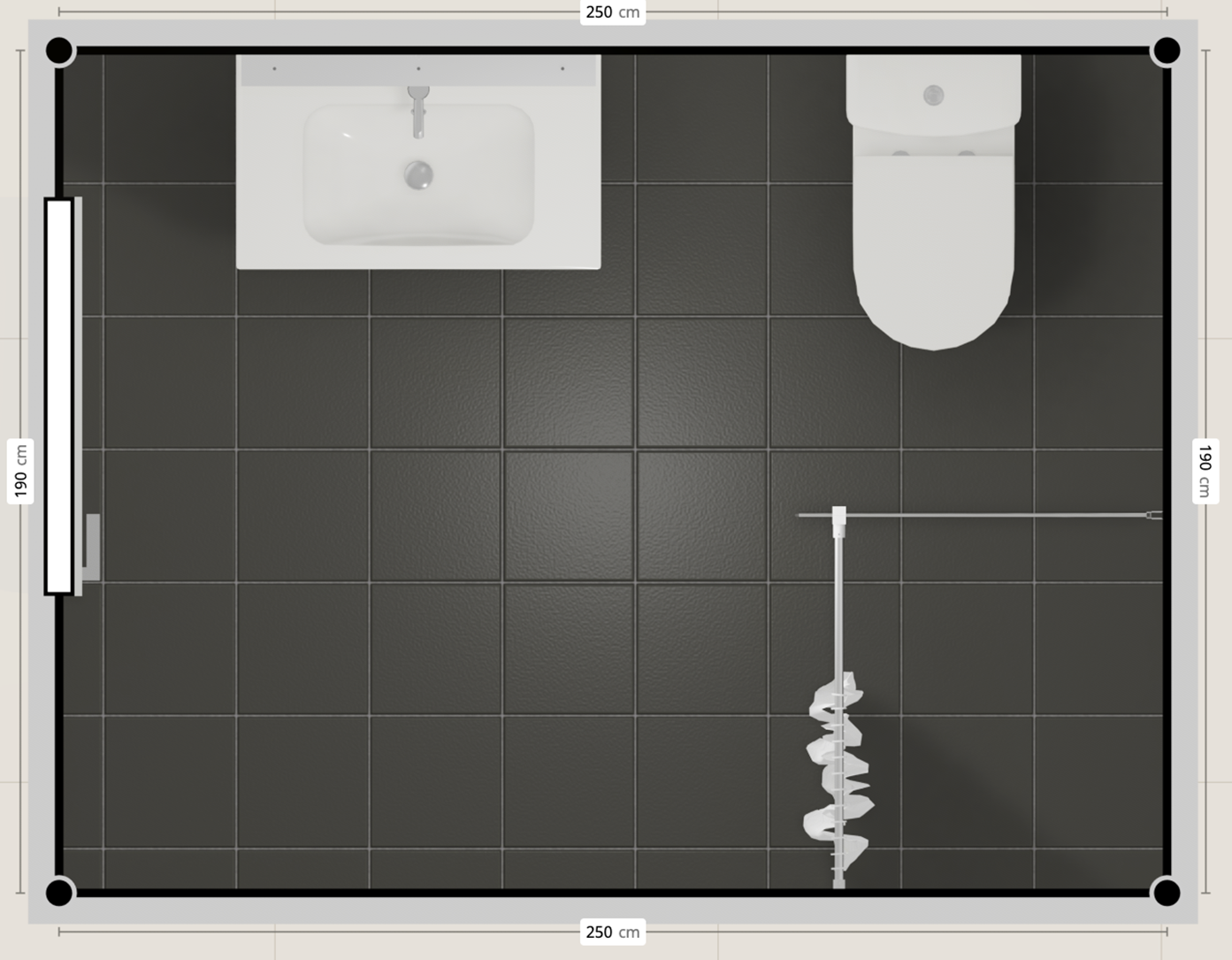}
  \caption{Top-down floor plan of the experimental bathroom (2.50\,m × 1.90\,m), showing the mmWave radar and vibration sensor mounting positions, annotated coverage areas, and key dimensions.}
  \label{fig:Floorplan}
  \vspace{-6mm}
\end{figure*}

\begin{figure*}[tb]
    \centering
    \includegraphics[width=1\linewidth]{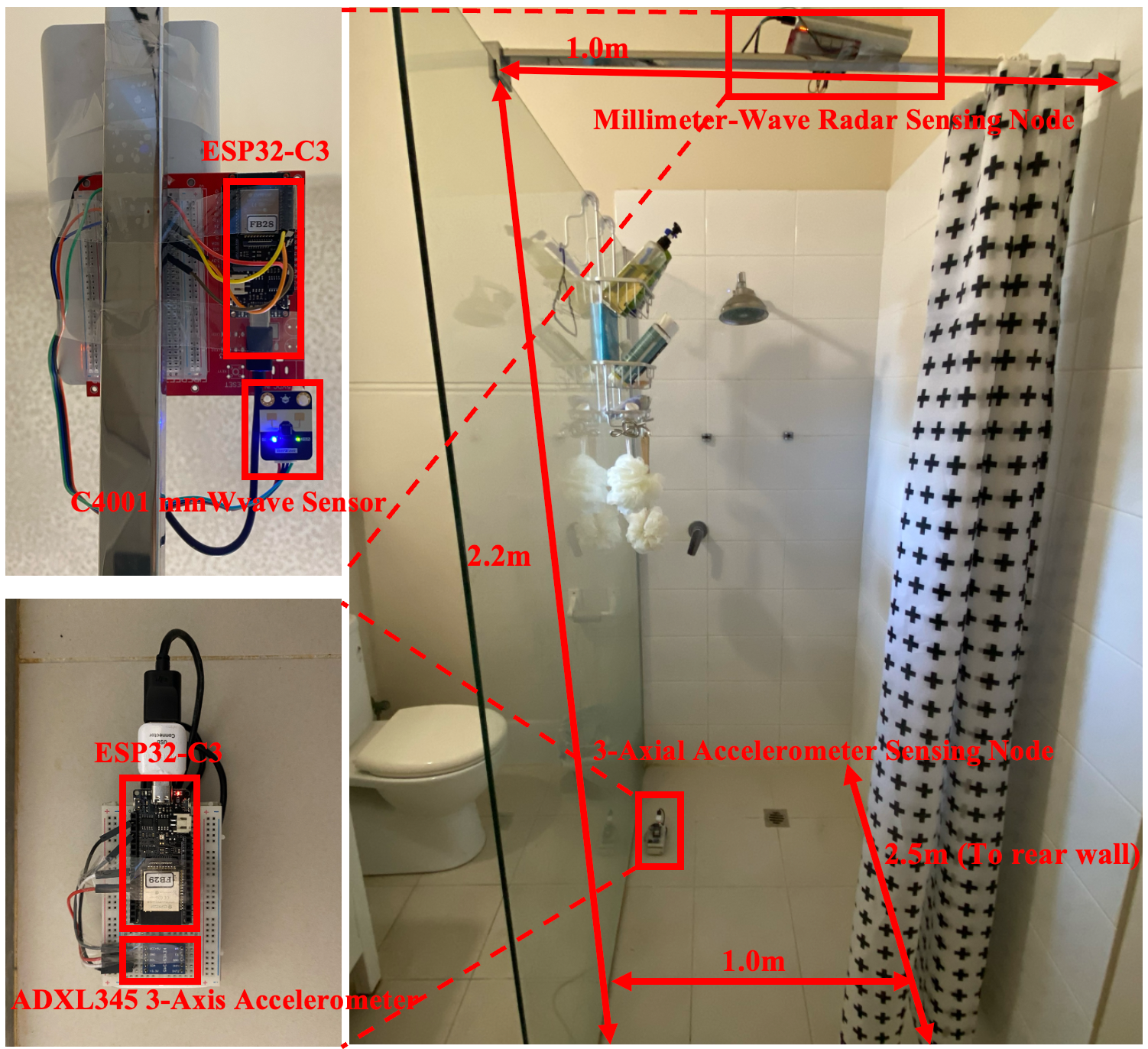}
    \vspace{-6mm}
    \caption{Illustration of the controlled bathroom environment and sensor placements. The mmWave radar node is mounted at 2.2\,m for broad coverage, while the ADXL345 vibration sensor is installed near the shower area to capture ground impacts. This setup closely replicates typical residential bathroom conditions for realistic fall detection experiments.}
    \label{fig:Exp_Setting}
    \vspace{-6mm}
\end{figure*}

To ensure the reliability, reproducibility, and real-world applicability of our proposed fall detection system, we conducted experiments in a controlled environment designed to replicate a typical residential bathroom (Fig.~\ref{fig:Floorplan} shows the Top-down floor plan). The experimental space measure 2.5m (L) × 1.1m (W) × 2.2m (H), forming a 6.05m² enclosed testing area. The bathroom features ceramic tiled walls, along with a glass partition and a shower curtain that resemble a typical residential bathroom. The floor consists of anti-slip ceramic tiles, with standard bathroom utilities including a showerhead, drain, and storage shelves containing toiletries such as soap and shampoo (as shown in Fig.~\ref{fig:Exp_Setting}).

\begin{figure*}[!b]
    \vspace{-6mm}
    \centering
    \includegraphics[width=1\linewidth]{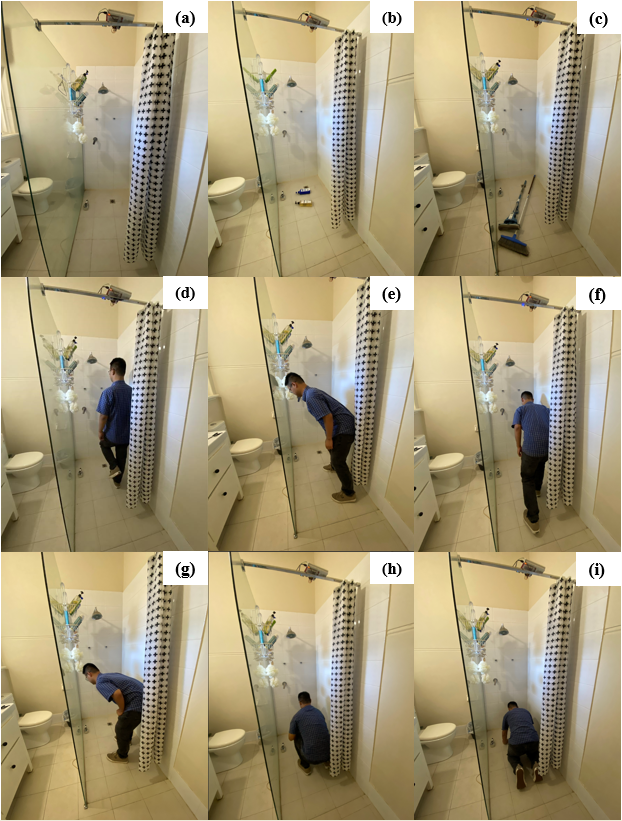}
    \vspace{-8mm}
    \caption{Overview of the eight experimental scenarios simulating typical bathroom conditions:(a) empty bathroom (no motion),(b) lightweight object drop,(c) heavy object drop,(d) normal walking,(e) bent-posture walking,(f) assisted walking with wall support,(g) static standing or squatting,(h)squat down and (i) fall down. These scenarios collectively capture a wide range of non-fall and fall-like events for robust real-world detection analysis.}
    \label{fig:Exp_behavior}
    \vspace{-8mm}
\end{figure*}

To achieve optimal sensor coverage while minimizing occlusion and environmental interference, two sensor nodes were strategically deployed within the test area. The triaxial vibration sensor node comprising an ADXL345 3-axis vibration, an ESP32-C3 microcontroller, and a rechargeable battery was securely placed 0.5m from the wall beneath the showerhead, a location frequently occupied during bathing activities. A waterproof case was used to protect the sensor from moisture exposure. The mmWave radar node consisting of a C4001 mmWave radar sensor, an ESP32-C3 MCU, and a battery unit was mounted on a metal beam at a height of 2.2m, ensuring unobstructed coverage of the entire bathroom space.

\vspace{-2mm}
\subsection{Experimental Design}
\label{sec:Experimental_Design}
\vspace{-2mm}

The evaluation protocol comprises nine representative bathroom scenarios (Fig.~\ref{fig:Exp_behavior}) chosen to span a spectrum of non‐fall and fall events, namely: (1) empty room (no motion), (2) light object drop (e.g., soap), (3) heavy object drop (e.g., mop), (4) normal walking, (5) bent‐posture walking, (6) wall‐supported walking, (7) static standing, (8) squatting, and (9) intentional falls. These scenarios were selected to challenge the system with variations in motion intensity, impact signature, and postural change, thereby testing P2MFDS’s ability to distinguish true falls from everyday disturbances and fall‐like activities.

Each scenario was conducted in multiple trials by all participants, resulting in a dataset comprising 20 minutes per scenario, with a total duration of 3 hours. The mmWave radar operated at a sampling rate of 10 Hz, generating 3D point clouds at 10 frames per second, capturing motion dynamics over time. Simultaneously, the triaxial vibration recorded 3D vibration data at 100 Hz, ensuring high temporal resolution for detecting impact forces associated with falls. The dataset included over 120,000 vibration data points and 18,000 mmWave frames, allowing for robust multimodal feature extraction. 

This experimental design ensures the system can effectively distinguish genuine falls from environmental disturbances, such as object drops, while accounting for variations in human movement patterns. 

\vspace{-2mm}
\subsection{Experimental Results}
\label{sec:Experimental_Results}
\vspace{-2mm}

\begin{figure*}[!t]
    \vspace{-4mm}
    \centering
    \includegraphics[width=1\linewidth]{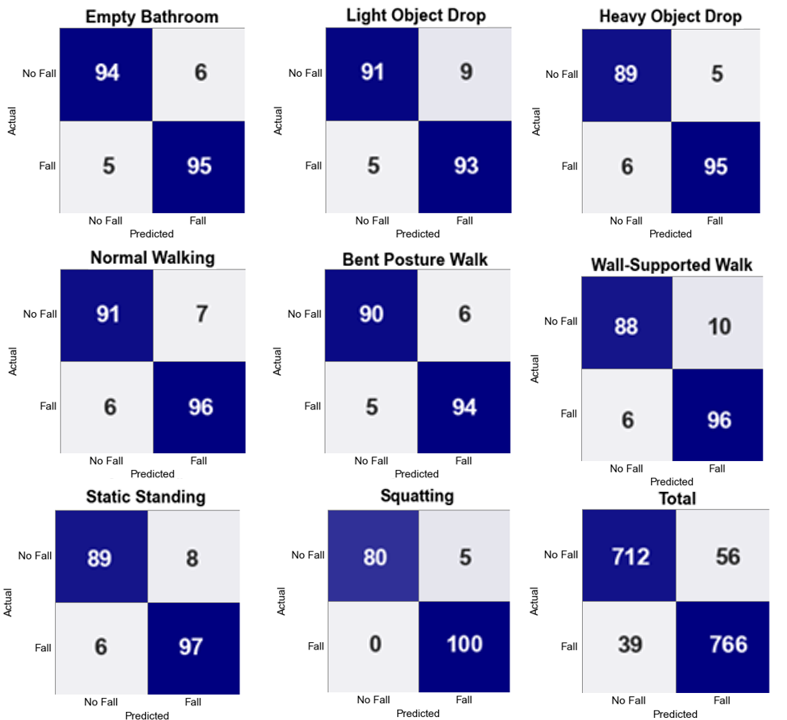}
    \vspace{-8mm}
    \caption{ Confusion matrices of the P2MFDS Network across eight experimental scenarios and the overall total. Each matrix shows classification counts for fall and non-fall events, illustrating high true positive and true negative rates with minimal misclassification.}
    \label{fig:confusion matrix}
    \vspace{-8mm}
\end{figure*}

The P2MFDS Network achieves an overall accuracy of 95.0\%, precision of 94.6\%, recall of 87.8\%, and F1‐score of 91.3\% across eight representative bathroom scenarios (Table~\ref{tab:cnn_lstm_conditions}). Scenario‐wise F1‐scores exceed 90\% in six activities, peaking at 97.9\% for normal walking and 97.3\% for squatting. The confusion matrices in Fig.~\ref{fig:confusion matrix} confirm a balanced distribution of true positives and true negatives, with squatting yielding zero false negatives and static standing maintaining a 93.9\% F1‐score despite potential environmental noise.

\begin{table}[!b]
\vspace{-6mm}
\caption{
P2MFDS performance across eight bathroom scenarios, reporting accuracy (Acc), recall (Rec), precision (Pre), and F1-score (F1) for non-fall and fall detection.
}
\label{tab:cnn_lstm_conditions}
\vspace{-2mm}
\centering
\setlength{\tabcolsep}{3.8pt}
\begin{tabular}{lcccccccc}
\toprule
\multirow{2}{*}{\textbf{Scenario}} 
& \multicolumn{4}{c}{\textbf{Non-Fall (\%)}} 
& \multicolumn{4}{c}{\textbf{Fall (\%)}} \\
\cmidrule(lr){2-5} \cmidrule(lr){6-9}
& Acc & Rec & Pre & F1 & Acc & Rec & Pre & F1 \\
\midrule
Empty Bathroom     & 96.0 & 96.1 & 96.1 & 96.1 & 96.0 & 95.9 & 95.9 & 95.9 \\
Light Object Drop  & 95.0 & 94.3 & 96.2 & 95.2 & 95.0 & 95.7 & 93.8 & 94.7 \\
Heavy Object Drop  & 90.4 & 91.4 & 89.8 & 90.6 & 90.4 & 89.5 & 91.1 & 90.1 \\
Normal Walking     & 94.7 & 95.7 & 93.8 & 94.7 & 94.7 & 93.8 & 95.7 & 94.8 \\
Bent Posture Walk  & 89.2 & 81.8 & 88.2 & 84.9 & 89.2 & 93.6 & 89.7 & 91.6 \\
Wall-Supported Walk& 85.3 & 89.1 & 84.5 & 86.7 & 85.3 & 80.9 & 86.4 & 83.5 \\
Static Standing    & 92.6 & 94.6 & 92.9 & 93.7 & 92.6 & 90.0 & 92.3 & 91.1 \\
Squatting          & 97.0 & 96.2 & 98.1 & 97.1 & 97.0 & 97.9 & 95.8 & 96.8 \\
\midrule
\textbf{Total}     & \textbf{95.0} & \textbf{98.0} & \textbf{95.1} & \textbf{96.5} 
                   & \textbf{95.0} & \textbf{94.6} & \textbf{87.8} & \textbf{91.3} \\
\bottomrule
\end{tabular}
\vspace{-6mm}
\end{table}

By fusing macro‐scale motion features from the mmWave radar with micro‐impact signatures from 3D vibration sensing, P2MFDS effectively suppresses false alarms in non‐fall events (e.g., 97.5\% F1 in an empty bathroom) while retaining high fall detection recall under complex disturbances (e.g., 94.6\% recall in light object drops). This robust performance across both quantitative metrics and confusion‐matrix analyses underscores the method’s resilience to noise and its practical viability for privacy‐preserving fall monitoring in confined bathroom environments.

\begin{table}[!t]
\vspace{-6mm}
\caption{
Comparison of P2MFDS with state-of-the-art methods on accuracy (Acc), precision (Pre), recall (Rec), and F1-score (F1).
}
\label{tab:performance_comparison}
\vspace{-2mm}
\centering
\setlength{\tabcolsep}{3.8pt}
\begin{tabular}{l l c c c c}
\toprule
\textbf{Methods} & \textbf{Model} & Acc (\%) & Pre (\%) & Rec (\%) & F1 (\%) \\  
\midrule
Rezaei et al. \cite{10048775} & CNN & 88.7 & 69.9 & 85.5 & 76.9 \\  
Maitre et al. \cite{9212552} & CNN-LSTM & 77.0 & 86.0 & 89.0 & 87.5 \\  
Yao et al. \cite{9776496} & RVAM Classifier & 86.7 & 84.7 & 88.9 & 86.8 \\  
Li et al. \cite{9747153} & CNN-LSTM & \textbf{96.8} & 84.6 & 89.6 & 87.0 \\  
He et al. \cite{10058501} & RBF NN & 85.9 & 88.3 & 82.8 & 85.4 \\  
Clemente et al. \cite{8678752} & SVM & 91.2 & 74.2 & 73.7 & 73.9 \\  
Hanif et al. \cite{9429253} & SVM & 89.0 & 88.0 & 87.0 & 87.5 \\
Wang et al. \cite{wang2023noncontact} & Transformer & 94.5 & 94.2 & 86.7 & 90.3 \\
Sadreazami et al. \cite{sadreazami2022compressed} & CNN & 82.9 & 92.8 & 81.2 & 86.6 \\
Swarubini et al. \cite{swarubini2024radar} & ResNet-50 & 67.2 & 50.6 & \textbf{98.8} & 66.8 \\
Yang et al. \cite{yang2022intelligent} & Bi-LSTM+CNN & 93.1 & 93.8 & 91.3 & 92.5 \\
Dai et al. \cite{dai2023multimodal} & YOLOX & 93.4 & 94.6 & 95.6 & 95.1 \\
Sun et al. \cite{sun2021attention} & Att-GRU & 94.2 & 92.4 & 83.8 & 87.9 \\
Meng et al. \cite{meng2022multimodal} & CNN-LSTM-Att & 93.1 & 94.3 & 91.2 & 92.7 \\
Alkhaldi et al. \cite{alkhaldi2022fall} & ResNet-18 & 93.0 & 91.4 & 90.2 & 90.8 \\
Zhang et al. \cite{zhang2023privacy} & Att-CNN-LSTM & 93.5 & 90.2 & 86.7 & 88.4 \\
\textbf{Ours} & \textbf{P2MFDS} & \underline{95.0} & \underline{\textbf{94.6}} & \underline{87.8} & \underline{\textbf{91.1}} \\  
\bottomrule
\end{tabular}
\vspace{-8mm}
\end{table}

\begin{table}[!b]
\vspace{-6mm}
\caption{
Ablation study showing the impact of model components (Attention, CNN, LSTM) and sensor modalities (Vibration, Radar) on P2MFDS performance (Acc, Rec, Pre, F1, AUC-ROC).
}
\label{tab:ablation_fall_detection}
\vspace{-2mm}
\centering
\renewcommand{\arraystretch}{1.2}
\setlength{\tabcolsep}{5pt}
\begin{tabular}{cccccccccc}
\toprule
\multicolumn{3}{c}{\textbf{Model Components}} & \multicolumn{2}{c}{\textbf{Sensors}} & \textbf{Acc} & \textbf{Rec} & \textbf{Pre} & \textbf{F1} & \textbf{AUC} \\
\hline
Att & CNN+LSTM & CNN & Vib & Rad & (\%) & (\%) & (\%) & (\%) & (\%) \\
\hline
\ding{51} & \ding{51} &        & \ding{51} &        & 92.3 & 80.6 & 83.4 & 82.0 & 86.2 \\
\ding{51} & \ding{51} &        &        & \ding{51} & 87.2 & 88.8 & 64.3 & 74.7 & 75.8 \\
\ding{51} & \ding{51} &        & \ding{51} & \ding{51} & 98.3 & 76.7 & 96.4 & 85.4 & 97.5 \\
\hline
\ding{51} &        & \ding{51} & \ding{51} &        & 84.5 & 51.2 & 70.0 & 59.1 & 72.1 \\
\ding{51} &        & \ding{51} &        & \ding{51} & 80.2 & 84.8 & 66.3 & 74.5 & 68.5 \\
\ding{51} &        & \ding{51} & \ding{51} & \ding{51} & 93.4 & 83.1 & 91.4 & 87.1 & 87.4 \\
\hline
\ding{51} & \ding{51} & \ding{51} & \ding{51} &        & 95.9 & 91.4 & 93.7 & 92.5 & 88.3 \\
\ding{51} & \ding{51} & \ding{51} &        & \ding{51} & 87.2 & 89.3 & 77.7 & 83.2 & 78.3 \\
\ding{51} & \ding{51} & \ding{51} & \ding{51} & \ding{51} & \textbf{95.0} & \textbf{94.6} & \textbf{87.8} & \textbf{91.1} & \textbf{96.4} \\
\bottomrule
\end{tabular}
\vspace{-8mm}
\end{table}

\vspace{-4mm}
\subsection{Comparison}
\label{sec:Comparison}
\vspace{-2mm}

To validate the effectiveness of our proposed Privacy-Preserving Multimodal Fall Detection System (P2MFDS) Network, we conducted a comprehensive performance comparison with 16 state-of-the-art fall detection models, as summarized in Table~\ref{tab:performance_comparison}. These methods span a diverse range of sensing modalities—including mmWave radar, UWB radar, Wi-Fi CSI, vibration sensors, and multimodal fusion—and model architectures such as CNNs, LSTMs, attention mechanisms, and SVMs.

Among radar-only methods, Rezaei et al.~\cite{10048775}, He et al.~\cite{10058501}, and Yao et al.~\cite{9776496} employed traditional CNN and RBF-based models, achieving moderate recall (82.8\%–88.9\%) but relatively lower precision, particularly in cluttered environments. Transformer-based~\cite{wang2023noncontact} and Bi-LSTM+CNN~\cite{yang2022intelligent} architectures showed improved performance due to better temporal modeling, yet still relied on a single sensing modality.

Several UWB radar-based solutions~\cite{9212552,sadreazami2022compressed,swarubini2024radar} demonstrated high recall (up to 98.8\%) but suffered from limited precision or overall accuracy, likely due to lower spatial resolution and vulnerability to occlusion. Similarly, vision-assisted fusion~\cite{dai2023multimodal} and Wi-Fi CSI-based fusion~\cite{meng2022multimodal} achieved high recall but may raise privacy concerns in sensitive environments like bathrooms.

Compared to these approaches, P2MFDS achieves competitive accuracy (95.0\%), superior precision (94.6\%), and robust recall (87.8\%) through effective multimodal integration of mmWave radar and 3D vibration sensing. This fusion enables the model to simultaneously capture macro-scale motion dynamics and micro-scale impact signals, enhancing robustness against noise and confounding activities. Notably, our method outperforms other multimodal fusion networks such as Sun et al.~\cite{sun2021attention} and Zhang et al.~\cite{zhang2023privacy}, both in terms of accuracy and precision. These results underscore the superiority of the P2MFDS architecture in balancing detection performance with privacy preservation in challenging, confined environments such as residential bathrooms.

\vspace{-4mm}
\subsection{Ablation Study}
\label{sec:Ablation_Study}
\vspace{-2mm}

To assess the impact of individual components in the Privacy-Preserving Multimodal Fall Detection System (P2MFDS) Network, we conducted an ablation study, as summarized in Table~\ref{tab:ablation_fall_detection}. The results demonstrate that multimodal sensor fusion significantly improves fall detection performance. Using only the vibration, the CNN+LSTM model achieves 92.3\% accuracy, while mmWave radar alone yields 87.2\% accuracy. When both modalities are combined, accuracy increases to 98.3\%, confirming the advantages of multimodal integration.

In terms of model architecture, the full P2MFDS Network (CNN+LSTM+Attention) achieves the highest performance, outperforming CNN- or LSTM-only configurations. The results highlight the complementary strengths of CNN for spatial feature extraction, LSTM for temporal modeling, and Attention for adaptive weighting, leading to robust and privacy-preserving fall detection.

\vspace{-4mm}
\section{CONCLUSION}
\label{sec:CONCLUSION}
\vspace{-4mm}

Our experimental results confirm that the Privacy-Preserving Multimodal Fall Detection System (P2MFDS) Network effectively fuses mmWave 3D point cloud and vibration data, achieving high accuracy under various activities such as object drops, squatting, and bent-posture walking. This fusion addresses limitations found in unimodal systems—particularly noise susceptibility and poor recall in cluttered bathroom settings. Furthermore, the attention mechanisms within the CNN-BiLSTM-Attention and Multi-Scale CNN-SEBlock-Self-Attention modules enhance the model’s adaptability to different motion patterns while preserving user privacy.

Despite these promising outcomes, environmental factors—such as temperature fluctuations or partial occlusions—may still pose challenges. Additionally, large-scale, long-term deployments in real homes with elderly  people are needed to further validate robustness and usability. Future work will focus on exploring adaptive calibration, extending the dataset across diverse bathroom layouts, and improving noise reduction to strengthen overall fall detection performance.

\vspace{-4mm}
\section{ACKNOWLEDGEMENT}
\label{sec:ACKNOWLEDGEMENT}
\vspace{-2mm}

This research was supported by the National Natural Science Foundation of China (Nos. 62172336 and 62032018). 

\vspace{-2mm}
\bibliographystyle{splncs04}
\bibliography{Fall}

\end{document}